\documentclass[10pt,twocolumn,letterpaper]{article}

\usepackage[pagenumbers]{cvpr} 

\usepackage[dvipsnames]{xcolor}

\usepackage{amsmath, amssymb}
\usepackage{graphicx}
\usepackage{subcaption}
\usepackage{booktabs}
\usepackage{enumitem}
\usepackage{lineno}

\usepackage[numbers,sort&compress]{natbib}

\definecolor{cvprblue}{rgb}{0.21,0.49,0.74}
\usepackage[pagebackref,breaklinks,colorlinks,citecolor=cvprblue]{hyperref}
\usepackage{array, booktabs, multirow, pifont}
\def\paperID{17} 
\def\confName{CVPR}
\def\confYear{2024}

\title{DSNet: A Novel Way to Use Atrous Convolutions in Semantic Segmentation}

\author{Zilu Guo$^{*1,2}$ \quad Liuyang Bian$^{1,2}$ \quad Xuan Huang$^{\dag2}$ \quad Hu Wei$^{2}$ \quad Jingyu Li$^{3}$ \quad Huasheng Ni$^{2}$ \\
\small $^{*}$First Author \qquad $^{\dag}$ Corresponding Author \vspace{1.8mm}\\
$^{1}$Anhui University, Institutes of Physical Science and Information Technology\\
$^{2}$Hefei Institutes of Physical Science, Chinese Academy of Sciences\\
$^{3}$University of Science and Technology of China\\
}

\begin{document}
\maketitle
\begin{abstract}
Atrous convolutions are employed as a method to increase the receptive field in semantic segmentation tasks. However, in previous works of  semantic segmentation, it was rarely employed in the shallow layers of the model. We revisit the design of atrous convolutions in modern convolutional neural networks (CNNs), and demonstrate that the concept of using large kernels to apply atrous convolutions could be a more powerful paradigm. We propose three guidelines to apply atrous convolutions more efficiently. Following these guidelines, we propose DSNet, a Dual-Branch CNN architecture, which incorporates atrous convolutions in the shallow layers of the model architecture, as well as pretraining the nearly entire encoder on ImageNet to achieve better performance. To demonstrate the effectiveness of our approach, our models achieve a new state-of-the-art trade-off between accuracy and speed on ADE20K, Cityscapes and BDD datasets. Specifically, DSNet achieves 40.0\% mIOU  with inference speed of 179.2 FPS on ADE20K, and 80.4\% mIOU with speed of 81.9 FPS on Cityscapes.  Source code and models are available at Github: https://github.com/takaniwa/DSNet.
\end{abstract}

\section{Introduction}
\label{sec:introduction}
 Semantic segmentation is a fundamental task in computer vision, requiring the prediction of each pixel in the input as a corresponding class \cite{firstsentence}. It finds wide applications in various fields such as autonomous driving, robot navigation, and medical image analysis\cite{autodr,ddrnet,her2net}.
\begin{figure}[h]
\centering
\includegraphics[width=8cm]{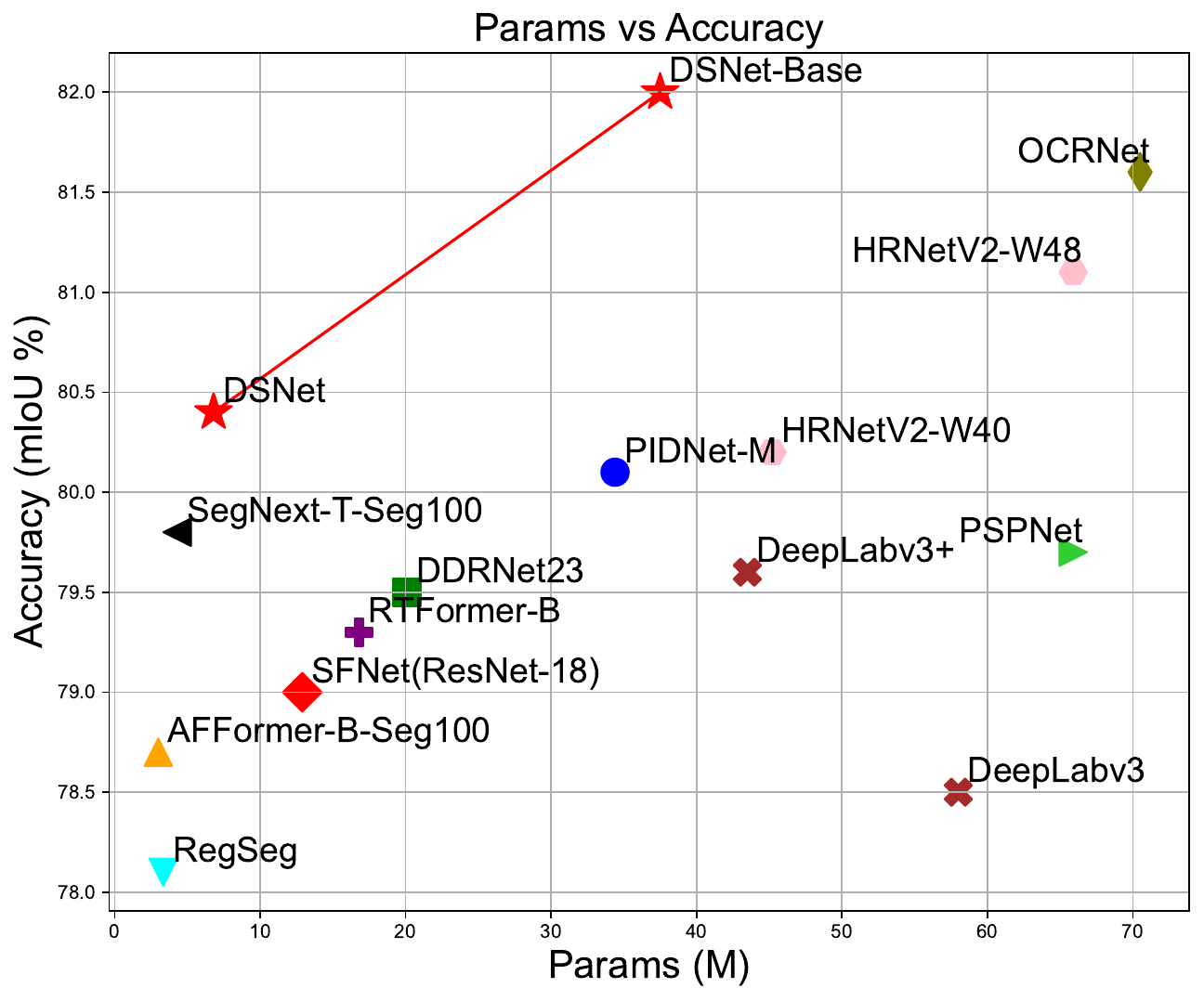}
\caption{Params vs mIOU on Cityscapes val set. Our model is depicted in red. We achieved a state-of-the-art balance between parameters and accuracy.}
\label{fig:paramvsacc}
\end{figure}

Recently, \textit{Convolutional Neural Networks(CNNs)} have encountered significant challenges in segmentation tasks from \textit{Vision Transformers(ViTs)}\cite{attentionisalltourneed,segformer}. Whether in high-precision segmentation tasks or real-time semantic segmentation tasks, \textit{ViTs} have demonstrated superior performance. RepLKNet\cite{LK} rethinks the factors contributing to the remarkable success of \textit{ViTs} in the field of computer vision. RepLKNet proposes that in \textit{ViTs}' architecture, the effective range of the receptive field\cite{erf} is more critical than the form of attention. Therefore, they present a novel approach using large kernels to enable the model to obtain a large effective receptive field, which leads to more similarity between representations obtained in shallow and deep layers. RepLKNet achieves performance comparable to or better than \textit{ViTs}.

RepLKNet\cite{LK} has inspired us to extend its design philosophy to atrous convolutions\cite{atrous}. In theory, convolutions can achieve a receptive field close to that of large kernels, making them a lightweight alternative solution. Similar to RepLKNet, stacking atrous convolutions in the shallow layers of the network may also produce promising results. However, in the supplementary experiments of the RepLKNet paper, atrous convolutions did not achieve the expected results as theorized. While many previous studies \cite{deeplabv1, deeplabv2, deeplabv3, regseg, hdcandduc, deeplabv3+} have focused on atrous convolutions, they either did not employ atrous convolutions in the shallow layers of the model or failed to fully leverage ImageNet pretraining to further improve performance. What factors contribute to this phenomenon? What effects might occur from stacking atrous convolutions starting from the shallow layers of the network?

To answer these questions, we rethought the design of atrous convolutions in \textit{CNNs}. Through a series of experiments, from a single-branch network to a dual-branch network, and from an atrous rate of 2 to an atrous rate of 15, we derived three empirical guidelines for atrous convolution: 
\begin{itemize}
    \item Do not only use atrous convolutions. Using atrous convolution along with dense convolutions may be a better choice.
    \item Avoiding the ``Atrous Disasters''. To achieve higher accuracy, selecting the appropriate atrous rate is crucial.
    \item Appropriate fusion mechanisms. Using an appropriate fusion mechanism to integrate information from different levels can improve the performance of the model.
\end{itemize}

Based on the above principles, we manually design a novel dual-branch network for semantic segmentation, Dual-branch with Same-resolution network(DSNet). This network demonstrates superiority in both real-time semantic segmentation and high-precision semantic segmentation. We also provide ablation experiments to demonstrate the functionality of each module. The main contributions of this paper are as follows:

\begin{itemize}
    \item  We revisited the design of atrous convolutions in \textit{CNNs}, and explored three empirical guidelines for atrous convolution. Based on the above guidelines, we proposed a novel Dual-branch network.
    \item DSNet achieved a new state-of-the-art trade-off between accuracy and speed on ADE20K, Cityscapes, and BDD. DSNet outperformed both real-time Transformer-based and convolutional neural network-based models on different datasets simultaneously.
\end{itemize}

\label{sec:1}
\section{Related Work}
\label{sec:relatedwork}
\subsection{High-Precision Semantic Segmentation}
\hspace{1.5em} In the early stages, semantic segmentation methods adopted encoder-decoder\cite{segnet} models, such as FCN\cite{fcn}, UNet\cite{unet}, SegNet\cite{segnet}. These models obtained high-level feature representations through continuous downsampling and then restored the resolution through upsampling or deconvolution. However, the output of the final layer of a regular encoder lacks spatial details and cannot be directly used for predicting segmentation masks. If only downsampling of the classification backbone is removed, the effective receptive field becomes too small to learn high-level semantic information. With the increasing demand for accuracy, many scholars began to focus on designing the model's encoder. DeeplabV3\cite{,deeplabv3} removed the last two stages of downsampling from the classification backbone and used atrous convolutions to establish long-range connections between pixels. HRNet\cite{hrnet} retained parallel branches with different resolutions in the backbone. RepLKNet\cite{LK} used large convolutional kernels to obtain an effective receptive field\cite{erf} and employed Structural Reparameterization\cite{repmlpnet,acnet} to address the issue of overly smooth large convolutional kernels that struggle to balance detail.
\begin{figure*}[h]
\centering
\includegraphics[width=16cm]{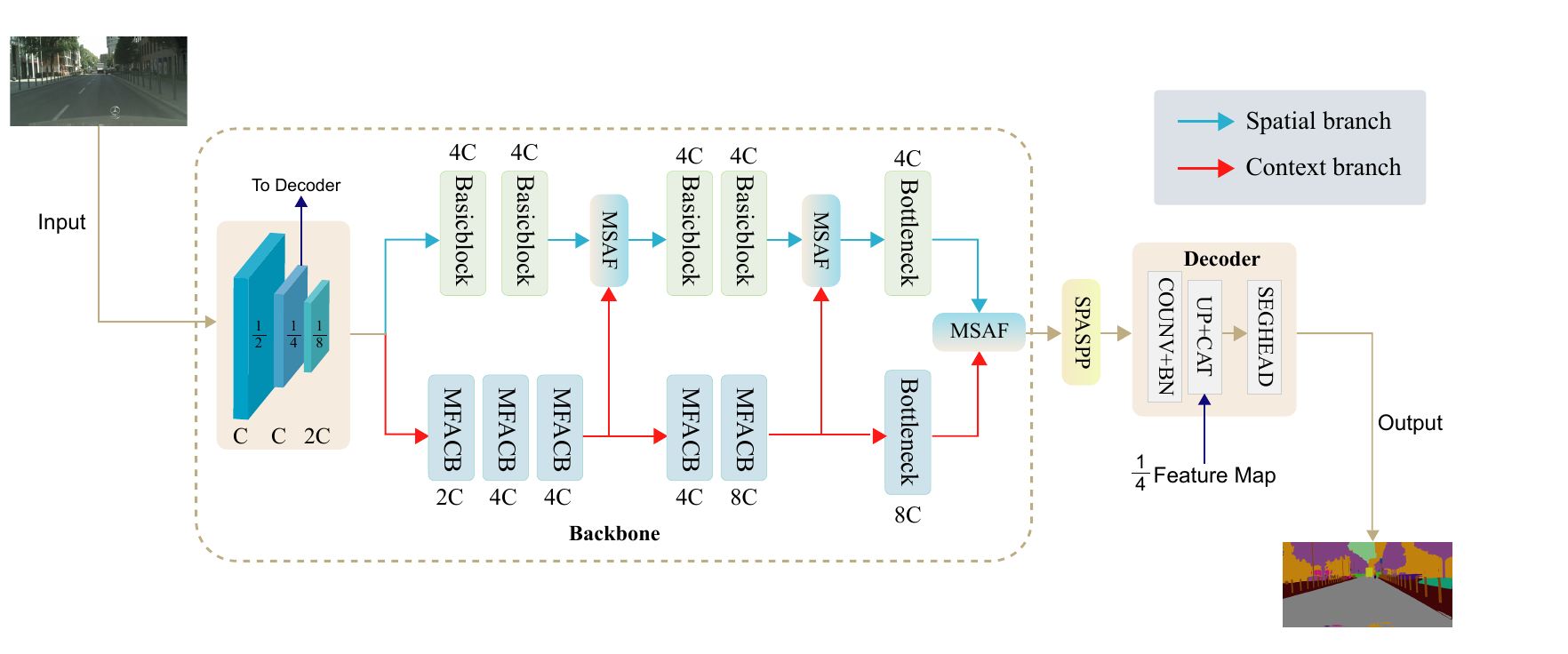}
\caption{Overview of DSNet. MFACB, MSAF, and SPASPP denotes Multi-scale Fusion
Atrous Convolutional Block, Multi-Scale Attention Fusion Module, and  Serial-Parallel Atrous Spatial Pyramid Pooling, respectively. UP indicates upsample, and CAT indicates Concatenate. C = 32.}
\label{model}
\end{figure*}
\subsection{Real-Time Semantic Segmentation}
\hspace{1.5em} BiseNet series\cite{bisenet,bisenetv2} use two paths (Spatial Path and Context Path) in the backbone and merge them at the end to achieve a balance between speed and accuracy. SFNet\cite{sfnet,sfnet-lite} delivers a Flow Alignment Module (FAM) to align feature maps of adjacent levels for better fusion. \cite{pidnet,ddrnet} follow the design philosophy of the BiseNet series. DDRNet\cite{ddrnet} employs a two-branch fusion with bilateral fusion and adds a context module at the end. PIDNet\cite{pidnet} proposes to expand the two branches into three branches: Spatial Path (P Path), Context Path (I Path), and Boundary Path (D Path) using the concept of a PID controller\cite{pidctr}. DDRNet and PIDNet are currently the best-performing real-time semantic segmentation models on Cityscapes\cite{cityscapes}. 

Recently, Many transformer-based lightweight segmentation methods have been proposed. TopFormer\cite{topformer} presents a new architecture that combines \textit{CNNs} and \textit{ViTs}. RTFormer\cite{rtformer} introduces a new network architecture that fully utilizes global context and improves semantic segmentation by deep attention without sacrificing efficiency. SeaFormer\cite{seaformer} introduces an attention mechanism leveraging both the squeeze-enhanced axial and the detail enhancement features, thereby architecting a novel framework termed the Squeeze-Enhanced Axial Transformer (SeaFormer), tailored for semantic segmentation in the mobile computing domain. TopFormer, RTFormer, and SeaFormer are currently the state-of-the-art real-time methods on ADE20K\cite{ade20k}.

\section{Method}
\label{method}

\subsection{Network design}
To answer the questions in Section \ref{sec:1}, we revisited the design of atrous convolutions in \textit{CNNs}, and summarized three empirical guidelines.

\noindent \textit{\textbf{Do not only use atrous convolutions.}} Many previous wor-
ks, such as DeepLab series\cite{deeplabv1,deeplabv2,deeplabv3} and Auto-DeepLab\cite{auto-deeplab}, were hesitant to use atrous convolutions in the shallow layers of the encoder. On the one hand, this may cause ``The Gridding Effect"\cite{hdcandduc,segformer}. Just like Large Kernel\cite{LK}, on the other hand, atrous convolution struggles to maintain a balance between contextual and detailed information, which can affect the model's performance. Inspired by BiseNet and DDRNet\cite{bisenet,ddrnet}, we propose to use atrous convolution along with dense convolution to maintain the detail information to eliminate the above disadvantages. From Table \ref{twotwo}, it can be observed that when using only atrous convolutions, the network's accuracy on both classification and segmentation tasks is relatively low. However, once the approach is adopted, the network's accuracy significantly improves.

\noindent\textit{\textbf{Avoiding the ``Atrous Disasters''.}} Despite demonstrating good performance in segmentation tasks, RegSeg\cite{regseg} heavily relies on atrous convolutions with an atrous rate of 14 in its network backbone. This limitation hinders its ability to achieve better initialization effects during ImageNet\cite{imagenet} pretraining. This hinders its further improvement in segmentation tasks. In the ImageNet classification task, the typical input image size is $224\times224$. Large atrous rates will result in the atrous convolution covering a range beyond the unpadded feature map and causing excessively large padding areas, this can limit the model's ability to learn better feature representations from ImageNet. From Table \ref{twotwo}, it can be observed that for atrous convolutions with a large atrous rate (e.g., d = 15), while they exhibit a slight advantage in segmentation tasks without pretraining on ImageNet, their segmentation performance significantly lags behind models using smaller atrous rates after pretraining on ImageNet. We refer to this phenomenon as the ``Atrous Disasters".
\begin{table}[h]
\centering
\renewcommand{\arraystretch}{1.1}
\setlength{\tabcolsep}{0.25mm}
\begin{tabular}{cccccc}
\hline
Atrous Rates & CB & SB & \begin{tabular}{c} Top1\\Acc.\\(\%) \end{tabular}  & \begin{tabular}{c}mIOU\\(w/o.p)\\(\%) \end{tabular}& \begin{tabular}{c}mIOU\\(w.p)\\(\%) \end{tabular}\\
\hline
None& & \checkmark&-&71.9&-\\
\hline
$d_{2}\times6+d_{3}\times6+d_{5}\times4$& \checkmark &  &-&74.2&-\\
\hline
$d_{2}\times3+d_{3}\times3+d_{15}\times10$& \checkmark & \checkmark &71.2&\textbf{78.3}&78.7\\
\hline
$d_{2}\times3+d_{3}\times3+d_{12}\times10$& \checkmark & \checkmark &72.0&78.2&79.0\\
\hline
$d_{2}\times3+d_{3}\times3+d_{5}\times10$& \checkmark & \checkmark &73.0&77.9&80.0\\
\hline
$d_{2}\times6+d_{3}\times6+d_{5}\times4$& \checkmark & \checkmark &73.1&77.9&\textbf{80.4}\\
\hline
$d_{2}\times6+d_{4}\times6+d_{6}\times4$& \checkmark & \checkmark &73.1&77.8&80.1\\
\hline
\end{tabular}
\caption{Ablations on Cityscapes Val set. Notation: CB indicates the Context Branch, SB indicates the Spatial Branch. $d_{2}$ indicates atrous rate = 2. Acc indicates the Top1 accuracy on ImageNet, w/o.p indicates the accuracy on Cityscapes val without ImageNet pretraining, w.p indicates the accuracy on Cityscapes with ImageNet pretraining.}
\label{twotwo}
\end{table}
Surprisingly, even though atrous convolutions with an atrous rate of 12 can correspond pixel by pixel to feature map pixels downsampled 8 times from ImageNet, the ``Atrous Disasters" still occurs. We attribute this phenomenon to the excessively large padding region, which reduces the actual effective range of atrous convolution, thus affecting the pretraining effectiveness of models on ImageNet. 

Therefore, if aiming to achieve higher accuracy in semantic segmentation through pretraining on ImageNet, selecting the appropriate atrous rate is crucial. To further expand the receptive field in semantic segmentation, consider integrating a context module\cite{deeplabv3,pspnet,ddrnet} outside the backbone network, which provides greater flexibility in its utilization.

\noindent \textbf{\textit{Appropriate fusion mechanisms.}} 
Simply merging information from different levels through operations like element-wise addition or concatenation is inadequate. These operations only provide a fixed linear aggregation of feature maps and do not clearly determine whether this combination is suitable for specific objects. An appropriate fusion mechanism can effectively guide the fusion of the two branches, enhance information transfer between features at different levels, and improve the model's representation ability. Many previous studies\cite{AFF,pp-liteseg,pidnet,bisenetv2} have shown that better results can be achieved by using an appropriate fusion mechanism compared to simple element-wise addition and concatenation, we further prove this conclusion through experiments in Section \ref{sec:4.3}.
\label{sec3.1}

\subsection{DSNet: A novel Dual-Branch Network}
Following the three guidelines in Section \ref{sec3.1}, we manually designed a dual-branch model, the model architecture diagram is shown in Figure \ref{model}. 

\begin{itemize}
    \item We split the network into two branches, the spatial branch and the context branch. The context branch is primarily composed of atrous convolutions, implemented as MFACB in Section \ref{sec:mfacb}, while the spatial branch consists of $3\times3$ dense convolutions.
    \item To fully integrate the information from both branches, three horizontal connections are made between the spatial branch and the context branch using MSAF from Section \ref{sec:msaf}.
    \item The backbone of the network primarily utilizes atrous convolutions with small atrous rates, such as 2, 3, and 5, to mitigate the ``Atrous Disasters". Additionally, the backbone is pretrained on ImageNet to enhance feature representation capabilities.
    \item We propose a context module named SPASPP in Section \ref{sec:spaspp}  to be inserted outside the backbone in segmentation tasks to rapidly increase the receptive field.
\end{itemize}
\noindent As both branches have the same resolution, we refer to it as a \textbf{\textit{Dual-branch with Same-resolution Network}}(DSNet). We simply designed two versions: DSNet is the lightweight version with fast inference, and DSNet-Base with high accuracy. DSNet-Base is a deeper version with more channels compared to DSNet.

\subsection{MFACB: Learning of different scales.}
\begin{figure}[h]
\centering
\includegraphics[width=8.5cm]{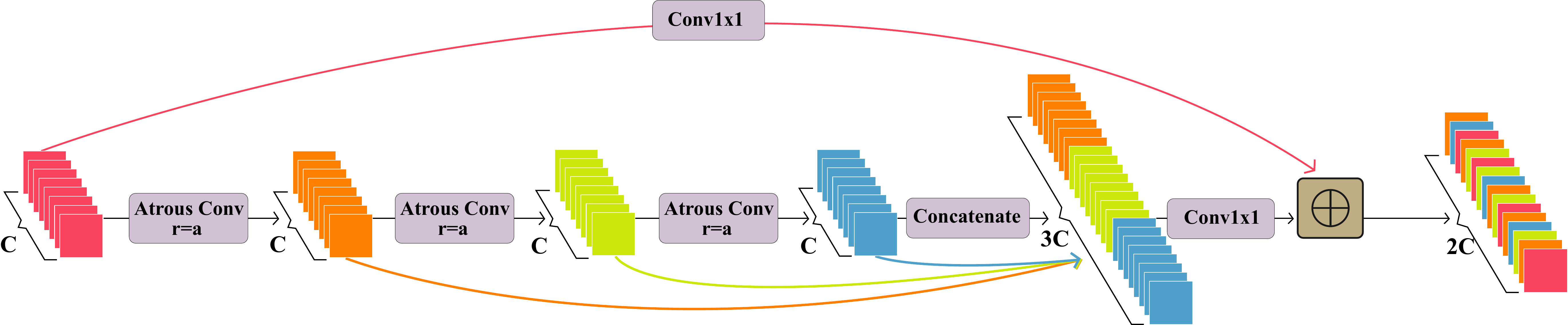}
\caption{Diagram of Multi-Scale Fusion Atrous Convolutional Block (MFACB). Where C represents the number of channels, and r = a indicates the atrous rate = a.}
\label{fig:mafcb}
\end{figure}

\hspace{1.5em} For better  perceptual abilities  at different scales, inspired by STDC \cite{STDC}, we introduce a novel encoder module for semantic segmentation, named the \textit{Multi-scale Fusion Atrous Convolutional Block} (MFACB). As shown in Figure \ref{fig:mafcb}, MFACB consists of three atrous convolutional layers, each using a different atrous rate to expand the receptive field. After three convolutional operations, the intermediate feature maps are concatenated and channel-wise compressed using a $1\times1$ convolution. Finally, the compressed feature maps are residual-connected with the input feature maps.

\begin{table}[h]
    \centering
    \renewcommand\arraystretch{1.1}
    \resizebox{\linewidth}{!}{\begin{tabular}{m{.08\linewidth}<{\centering}|m{1.12\linewidth}}
        \hline
        \multirow{8}{*}[-13ex]{RF}& \begin{tabular}{p{.27\linewidth}@{\quad}|p{.27\linewidth}@{\quad}|p{.27\linewidth}@{\quad}}
            MFACB1[\textbf{2},2,2] & MFACB1[\textbf{2,2},2] & MFACB1[\textbf{2,2,2}]
        \end{tabular} \\
        \cline{2-2}
         & \begin{tabular}{p{.27\linewidth}<{\centering}@{\quad}|p{.27\linewidth}<{\centering}@{\quad}|p{.27\linewidth}<{\centering}@{\quad}}
            \{ $5 \times 5$ \} & \{ $9 \times 9$ \} & \{ $13 \times 13$ \}
        \end{tabular} \\
        \cline{2-2}
         & \begin{tabular}{p{.45\linewidth}<{\centering}@{\quad}|p{.45\linewidth}<{\centering}@{\quad}}
            MFACB2[\textbf{2,2,2}], Cat & MFACB2[\textbf{2,2,2}], Fusion
        \end{tabular} \\
        \cline{2-2}
         & \begin{tabular}{p{.45\linewidth}<{\centering}@{\quad}|p{.45\linewidth}<{\centering}@{\quad}}
            $\left\{\begin{array}[c]{c}
                5 \times 5 \\
                9 \times 9 \\
                13 \times 13
            \end{array}\right.$ & $ a = \left\{\begin{array}[c]{c}
                5 \times 5 \\
                9 \times 9 \\
                13 \times 13
            \end{array}\right.$
        \end{tabular} \\
        \cline{2-2}
        & \begin{tabular}{p{.27\linewidth}<{\centering}@{\quad}|p{.27\linewidth}<{\centering}@{\quad}|p{.27\linewidth}<{\centering}@{\quad}}
           MFACB2[\textbf{3},3,3] & MFACB2[\textbf{3,3},3] & MFACB2[\textbf{3,3,3}]
        \end{tabular} \\
        \cline{2-2}
         & \begin{tabular}{p{.27\linewidth}<{\centering}@{\quad}|p{.27\linewidth}<{\centering}@{\quad}|p{.27\linewidth}<{\centering}@{\quad}}
            $b = \left\{\begin{array}[c]{c}
                11 \times 11 \\
                15 \times 15 \\
                19 \times 19
            \end{array}\right.$ & $ c = \left\{\begin{array}[c]{c}
                17 \times 17 \\
                21 \times 21 \\
                25 \times 25
            \end{array}\right.$ & $ d = \left\{\begin{array}[c]{c}
                23 \times 23 \\
                27 \times 27 \\
                31 \times 31
            \end{array}\right.$
        \end{tabular} \\
        \cline{2-2}
        & \begin{tabular}{p{.45\linewidth}<{\centering}@{\quad}|p{.45\linewidth}<{\centering}@{\quad}}
            MFACB2[\textbf{3,3,3}], Cat & MFACB2[\textbf{3,3,3}], Fusion
        \end{tabular} \\
        \cline{2-2}
         & \begin{tabular}{p{.45\linewidth}<{\centering}@{\quad}|p{.45\linewidth}<{\centering}@{\quad}}
            [\textbf{b, c, d}] & [\textbf{a}, b, c, d]
        \end{tabular} \\
        \hline
    \end{tabular}}

        \caption{Receptive Field of layer in our MFACB module. RF de-
notes Receptive Field.  }
    \label{rf_a}
\end{table}

We can gain deeper insights into the role of this module by observing Table \ref{rf_a}. After the first MFACB module with atrous rates of [2, 2, 2], the receptive fields of the three intermediate feature maps are $5\times5$, $9\times9$, and $13\times13$, respectively. Subsequently, through concatenation and $1\times1$ convolution, the output feature map aggregates these three different-scale receptive fields simultaneously. After the concatenation and fusion operations in the second MFACB module, for the same reason, the scale of the receptive field of the current layer begins to increase. Using MFACB in the backbone network allows the model to effectively learn semantic information at different scales.

\label{sec:mfacb}

\subsection{MSAF: Balancing the Details and Contexts}
Skip connections enhance the information transmission between feature maps at different layers or scales, thus improving the model's representational capacity. Follow the third suggestion in Section \ref{sec3.1}, we introduce a novel \textit{Multi-Scale Attention Fusion Module} (MSAF), aimed at enabling selective learning between two different-level branches without overwhelming them. The main idea is to let the network learn feature weights based on the loss, allowing the model to selectively fuse information from different scales. This module can be mainly divided into two parts: \textit{Multi-Scale Attention}(MSA) and\textit{ Multi-Scale Attention Fusion} Module(MSAF).

\begin{figure}[h]
\centering
\includegraphics[width=8.5cm]{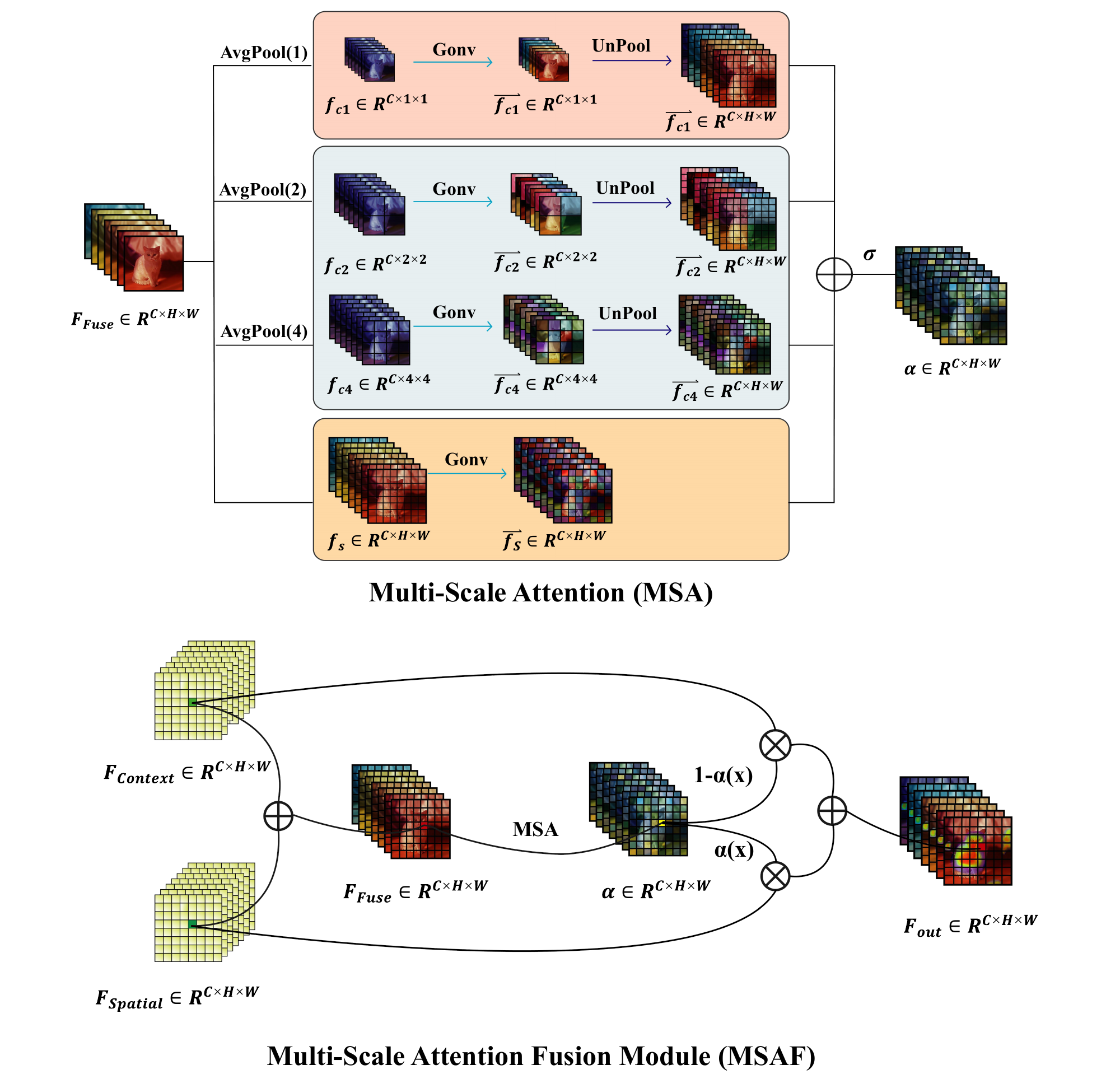}
\caption{MSA and MSAF schematic diagram. AvgPool(4) denotes global average pooling to $4\times4$, $\sigma$ represents the sigmoid function. Unpool represents average unpooling.}
\label{msaf}
\end{figure}

\noindent\textbf{\textit{Multi-Scale Attention (MSA). }}As shown in Figure \ref{msaf}, the main purpose of MSA is to learn the weights $\alpha$ as the basis for the fusion of different-level branches. The MSA
module is mainly divided into two parts: Region Attention and Pixel Attention. To provide a more detailed explanation of MSA, we denote the number of channels, feature map width, and height as C, W, and H, respectively.

1): \textit{Region Attention}: Region attention measures the importance of different regions in the feature map. We propose that considering the receptive field when using attention is necessary. For convenience, we divide the feature map into blocks of the same size, such as $1\times1$ for channel, $2\times2$ for 4 blocks, and $4\times4$ for 16 blocks. In DSNet, we divide the feature map into equally sized regions of $1\times1$, $4\times4$, $8\times8$, and $16\times16$. Taking $4\times4$ as an example, we first perform average pooling on the feature map $F_{Fuse} \in \mathbb{R}^{C \times H \times W}$ to obtain $f_{c4} \in \mathbb{R}^{C \times 4 \times 4}$, then perform channel compression and expansion to obtain $\overrightarrow{f_{c4}} \in \mathbb{R}^{C \times 4 \times 4}$. For compatibility with pixel attention, we reshape it back to $\overrightarrow{f_{c4}} \in \mathbb{R}^{C \times H \times W}$. The mathematical formulas of Region Attention can be described as Equation \ref{eq4}. Similar to SE attention\cite{senet}, the purpose of channel compression and expansion is to reduce computational complexity and enhance non-linearity. \textit{GONV} represents the operations of channel expansion and compression.
\begin{equation}
F_{Fuse} = F_{Context} + F_{Spatial}
\label{eq1}
\end{equation}
\begin{equation}
\overrightarrow{f_{reg}} = \sum_{i=1,4,8,16} \text{\textit{UnPool}}\left( \text{\textit{Gonv}}\left( \text{\textit{AvgPool}[{i}]} \left( F_{Fuse} \right) \right) \right)
\label{eq4}
\end{equation}

2): \textit{Pixel Attention}: Pixel attention measures the importance of each pixel. This module does not require pooling and reshaping. 
As shown in Equation \ref{eq3}, we directly perform channel compression and expansion on $F_{Fuse} \in \mathbb{R}^{C \times H \times W}$ to obtain $\overrightarrow{f_{S}} \in \mathbb{R}^{C \times H \times W}$. For same reason, $\overrightarrow{f_{S}}$ can measure the importance of each pixel.
\begin{equation}
\overrightarrow{f_{S}} = \text{\textit{Gonv}}\left( F_{Fuse} \right) \quad
\label{eq3}
\end{equation}

\noindent\textbf{\textit{Multi-Scale Attention Fusion Module.}} We obtain the weights of different positions in the feature map by overlaying pixel attention and region attention, and the formula for deriving the weights can be represented by Equation \ref{eq5}. As shown in the Figure \ref{msaf}, we finally fuse the two branches by element-wise multiplication. The mathematical formulas of region attention can be described as Equation \ref{eq6}.
\label{sec:msaf}
\begin{equation}
\alpha = \text{\textit{Sigmoid}}\left( \text{\textit{Add}}\left( \overrightarrow{f_{s}}, \overrightarrow{f_{reg}} \right) \right)
\label{eq5}
\end{equation}
\begin{equation}
F_{\text{Out}} = F_{\text{Context}} \times \alpha + F_{\text{Spatial}} \times (1 - \alpha)
\label{eq6}
\end{equation}

\subsection{SPASPP: Further extracting context information}
\hspace{1.5em} We propose a new module to further extract context information from feature maps. 
illustrates the internal structure of \textit{Serial-Parallel Atrous Spatial Pyramid Pooling}(SPASPP). Unlike the fully parallel structure of ASPP\cite{deeplabv3}, we stack several $3\times3$ atrous convolutions. 
Subsequently, we concatenate the intermediate feature maps obtained by stacking the atrous convolutions with the upsampled feature maps after global pooling. Stacking atrous convolutions and concatenating is intended to rapidly increase the receptive field using this context module outside the pretrained ImageNet backbone network, while also gaining context information at different scales. Then, we compress the channels using $1\times1$ convolution and perform residual connection with the input. 
\begin{figure}[h]
\centering
\includegraphics[width=6cm]{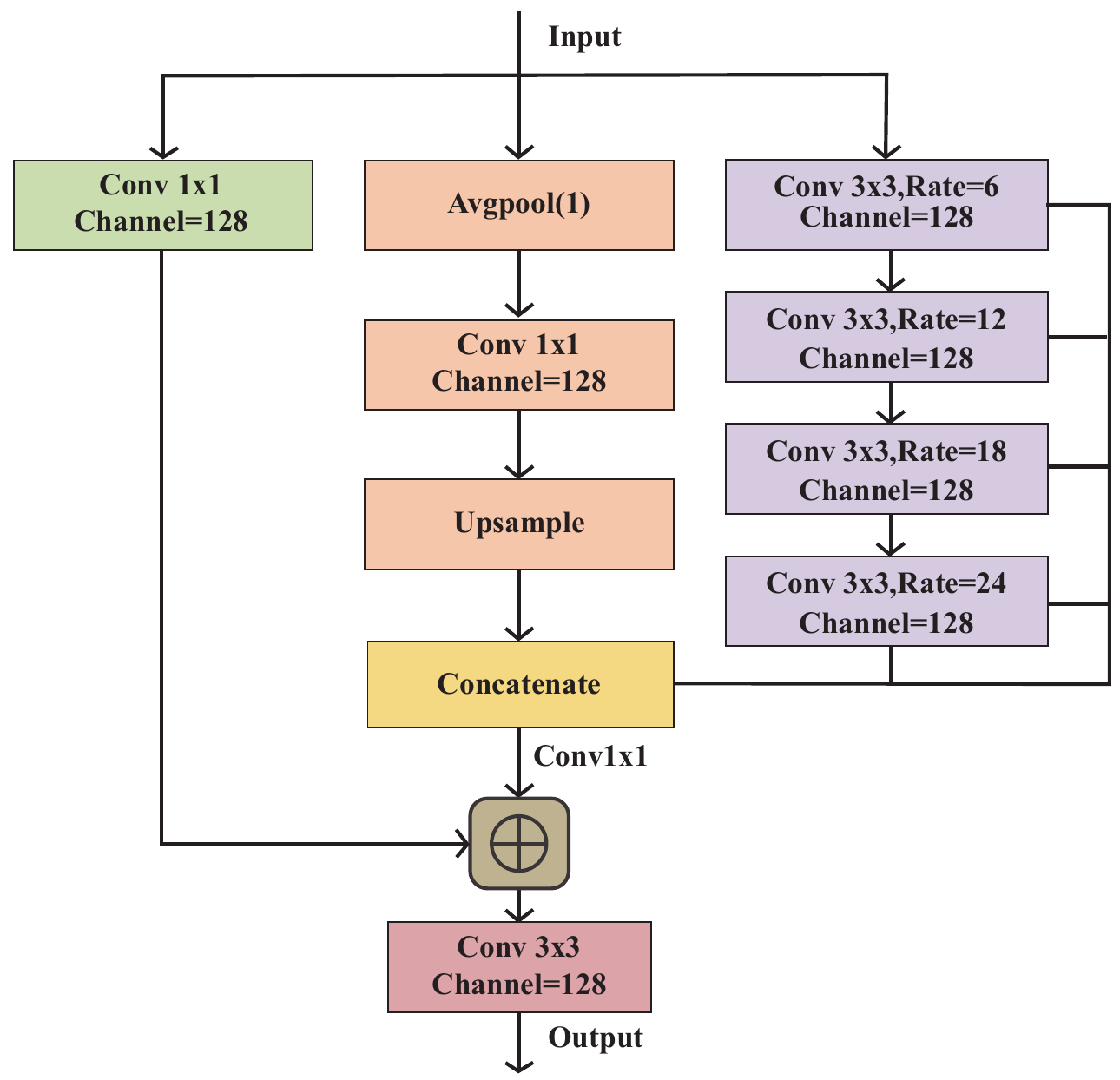}
\caption{Illustration of SPASPP module. }
\label{spaspp}
\end{figure}Figure \ref{spaspp}
The fusion of convolutional kernels with different atrous rates forms a multi-scale property. The purpose of using an additive mode rather than a fully parallel mode is to allow the model to obtain richer context information and further expand the receptive field, thus performing better in tasks requiring a large receptive field. The Table \ref{SPASPP} demonstrates that SPASPP can offer richer context information compared to ASPP without the addition of extra convolutional layers, thus maintaining a speed essentially equivalent to ASPP.
\label{sec:spaspp}
\begin{table}[h]
\centering
\setlength{\tabcolsep}{0.5mm}
\renewcommand{\arraystretch}{1.1}
\begin{tabular}{ccc}
\hline
 &ASPP&SPASPP\\
\hline
Context Scale&[6, 12, 18, 24, all]&\begin{tabular}{c}{[6, all]}\\{[6, 12, all]}\\{[6, 12, 18, all]}\\{[6, 12, 18, 24, all]}\\
\end{tabular}\\
\hline
\end{tabular}
\caption{Differences between SPASPP and ASPP.}
\label{SPASPP}
\end{table}
\section{Experiment}
\subsection{Dataset}
We perform segmentation experiments over ADE20K,
Cityscapes, and BDD. The ADE20K\cite{ade20k} dataset covers 150 categories and contains 25,000 images, which are split into 20,000 for training, 2,000 for validation, and 3,000 for testing. During testing, all images are resized to $512\times512$. Cityscapes\cite{cityscapes} is a publicly available resource designed for semantic segmentation tasks. It contains 2975 finely annotated images for training, 500 images for validation, and 1525 images for testing. The image resolution is $2048\times1024$, which is challenging for real-time models. Only the fine annotated dataset is used here. The BDD\cite{bdd100k} dataset is a comprehensive repository tailored for autonomous driving applications, featuring 19 distinct classes. It encompasses a training set of 7,000 images and a validation set of 1,000 images, each image sized at $1280\times720$ pixels.
\subsection{Implementation Details}
\textbf{Training.} After pretrained on ImageNet\cite{imagenet}, our training protocols on semantic segmentation tasks are almost the same as previous works\cite{ddrnet,regseg,pidnet,sfnet,sfnet-lite}. We use the SGD optimizer with the momentum of 0.9. As a common practice, the “poly” learning rate policy is adopted to decay the initial learning rate. Data augmentation contains random horizontal flip, random resizing with a scale range of [0.4, 1.6], and random cropping. For DSNet, the number of iterations, initial learning rate, weight decay, cropped size and batch size for Cityscapes, ADE20K, BDD could be summarized as [120k, 0.01, 0.0005, $1024\times1024$, 24], [150k, 0.02, 0.0001, $512\times512$, 32], [87k, 0.01, 0.0005, $512\times512$, 24], respectively. For DSNet-Base, the number of iterations, initial learning rate, weight decay, cropped size and batch size for Cityscapes, ADE20K, BDD could be summarized as [120k, 0.01, 0.0005, $1024\times1024$, 32], [160k, 0.02, 0.0001, $512\times512$, 32], [108k, 0.01, 0.0005, $512\times512$, 16], respectively.

\noindent \textbf{Inference.} We measure inference speed on a platform consisting of a single RTX 4090, PyTorch 1.10, CUDA 11.3, cuDNN 8.0, and an Ubuntu environment. Following\cite{ddrnet,pidnet}, we integrate batch normalization into the convolutional layers and set the batch size to 1 in order to measure the inference speed.

\noindent \textbf{Special speed comparison.} In particular, to facilitate speed comparison with some models developed based on the mmcv framework\cite{mmcv}, we additionally used the RTX3090 to infer the speed, since it is a significant time overhead to either port our model to the mmcv framework or to port the model based on the mmcv framework to our environment. The speed of the mmcv-based model can be found in SCTNet\cite{sctnet}.

\subsection{Ablation Study}

\noindent \textbf{MSAF for Two-branch Networks.} We applied different fusion methods to the three lateral connections of the model. The experimental results are shown in Table \ref{msafstudy}. Compared to element-wise addition(Add), all three attention fusion methods demonstrated superiority in the segmentation task, improving accuracy with minimal speed loss. Particularly, our proposed multi-scale fusion module exhibited superior performance in both classification and segmentation tasks, achieving a 0.9\% improvement on the Imagenet1k classification task and a 0.7\% improvement on Cityscapes. Compared to AFF and iAFF\cite{AFF}, it still achieved a nearly 0.4\% increase in accuracy. This experiment evaluated the third recommendation regarding the use of atrous convolutions proposed in Section \ref{sec3.1}.

\begin{table}[h]

    \centering
    \resizebox{\linewidth}{!}{\begin{tabular}{m{.1\linewidth}<{\centering}m{.05\linewidth}<{\centering}m{.05\linewidth}<{\centering}m{.07\linewidth}<{\centering}m{.11\linewidth}<{\centering}m{.16\linewidth}<{\centering}m{.15\linewidth}<{\centering}m{.1\linewidth}<{\centering}}
        \toprule
        \multirow{2}{*}{Model} & \multicolumn{4}{c}{Fusion} & \multirow{2}{*}{Top1 Acc.} & \multirow{2}{*}{mIOU(\%)} & \multirow{2}{*}{\#FPS} \\
        \cmidrule{2-5}
        & Add & AFF & iAFF & MSAF & & &  \\
        \midrule
        \multirow{4}{*}{DSNet} 
        & \ding{51} & & & & 72.3 & 79.7 & 90.5 \\ \cmidrule{2-8}
        &  & \ding{51} & & & 73.0 & 80.1 & 87.6 \\ \cmidrule{2-8}
        &  & & \ding{51} & & \textbf{-} & 80.0 & 81.1 \\ \cmidrule{2-8}
        &  & & & \ding{51} & \bfseries 73.2 & \bfseries 80.4 & 81.9 \\ 
        \bottomrule
    \end{tabular}}
    \caption{Performance of different fusion methods on cityscapes val set. AFF, iAFF stand for attentional feature fusion, iterative attentional feature fusion\cite{AFF}, respectively. }
    \label{msafstudy}
\end{table}

\noindent \textbf{Efficiency of MFACB.} To demonstrate the effectiveness of MFACB, we conducted a simple comparative experiment. We employed two different strategies, referred to as EXP 1 and EXP 2 in the Table \ref{MFACBEXCP}. To ensure fairness in the experiment, we employed the same training strategies, using two RTX 4090 GPUs, a batch size of 12, and 50,000 iterations for both experiments. From the table, it can be observed that with the same atrous rate and the same number of atrous convolutions, MFACB achieved better performance compared to BasicBlock\cite{resnet}. This further reinforces our determination to use MFACB in the Context Branch.
\begin{table}[h]

\centering
\setlength{\tabcolsep}{0.8mm}
\renewcommand{\arraystretch}{1.1}
\begin{tabular}{ccc}
\hline
 & Method & mIOU(\%)\\
\hline
 EXP 1 & \begin{tabular}{c}
        BasicBlock[ C=128, d=2] × 3\\
        BasicBlock[ C=256, d=4] × 3
        \end{tabular} 
    & 69.28 \\ 
\hline
 EXP 2 & \begin{tabular}{c}
        MFACB([64, 128, 128], d=2)\\
        MFACB([128, 128, 128], d=2)\\
        MFACB([128, 256, 256], d=4)\\
        MFACB([256, 256, 256], d=4)\\

        \end{tabular} 
    & \textbf{70.28} \\ 
\hline
\end{tabular}
\caption{Performance of MFACB on cityscapes val set.}
\label{MFACBEXCP}
\end{table}

\noindent \textbf{Efficiency of SPASPP.} We compared SPASPP with other context extraction modules, including Atrous Spatial Pyramid Pooling (ASPP)\cite{deeplabv3} and Deep Aggregation Pyramid Pooling Module (DAPPM)\cite{ddrnet}. To achieve a higher baseline, we did not reduce the intermediate channels of DAPPM. From Table \ref{SPSAPPEXP}, it can be seen that SPASPP proposed in this paper increased the accuracy from 77.3\% to 80.4\%, with an inference time increase of only 1-2ms. Furthermore, SPASPP improved the accuracy by 0.7\% over ASPP with almost no speed loss. 
\begin{table}[h]
\centering
\renewcommand{\arraystretch}{1.1}
\begin{tabular}{ccccc}
\hline
DAPPM & ASPP & SPASPP & mIOU(\%) &\#FPS\\
\hline
 {}& {} & {} & 77.3 & 91.8\\
 
 \checkmark&  &  & 79.3 & 80.2\\
 
& \checkmark &  & 79.7 & 82.3\\

&  &\checkmark & \textbf{80.4} & 81.9\\
\hline
\end{tabular}
\caption{Comparison of SPASPP and other context modules.}
\label{SPSAPPEXP}
\end{table}


\label{sec:4.3}

\subsection{Comparison}

\textbf{ADE20K.} Recently, most real-time CNN-based segmentation methods have emphasized mainly on Cityscapes, with little attention paid to the ADE20K dataset. A recent work\cite{sctnet} argued that ADE20K posed a considerable challenge to lightweight CNN-based models because it collects a large number of images spanning more than 150 categories. From Table \ref{Ade20k}, however, we have achieved good results on ADE20K, unlike other CNN-based models. The experimental results show that DSNets achieve the best trade-off between inference speed and accuracy. Specifically, DSNet achieves higher accuracy than SegFormerB0, TopFormer-B, and RTFormer-S, surpassing them by 2.6\%, 0.8\%, and 3.3\%, respectively. Additionally, DSNet operates at approximately twice the speed of these models. DSNet-Base also outperforms SeaFormer-B and SegNext-T in accuracy while maintaining a faster speed. Compared to RTFormer-B, DSNet-Base still maintains a 1.3\% accuracy advantage.
\begin{table}[h]
\centering
\setlength{\tabcolsep}{0.2mm}
\renewcommand{\arraystretch}{1.1}
\begin{tabular}{ccccc}
\hline Method &Reference& $\mathrm{mIOU}(\%)$ &\#FPS & GPU \\
\hline \textit{mmcv-based} & & & \\
SegFormerB0\cite{segformer}&NeurIPS21 & 37.4 & 84.4 & RTX3090 \\
SeaFormer-B\cite{seaformer}&ICLR23 & 41.0 & 44.5 & RTX3090 \\
TopFormer-B\cite{topformer}&CVPR22b & 39.2 & 96.2 & RTX3090 \\
SegNext-T\cite{segnext}&NeurIPS22b & 41.1 & 60.3 & RTX3090 \\
RTFormer-S\cite{rtformer}&NeurIPS22 & 36.7 & 95.2 & RTX3090 \\
RTFormer-B\cite{rtformer}&NeurIPS22 & 42.1 & 93.4 & RTX3090 \\
\hline DSNet-head256&Ours & 40.0 & $\mathbf{1 7 9 . 2}$ & RTX3090 \\
DSNet-Base &Ours & $\mathbf{4 3 . 4}$ & 66.8 & RTX3090 \\
\hline
\end{tabular}
\caption{Comparisons with other state-of-the-art real-time methods on ADE20K. The FPS is measured at resolution 512 × 512. All methods measured by single scale inference. Head256 denotes the number of output channels is 256, the same below.}
\label{Ade20k}
\end{table}

\noindent \textbf{BDD.}   For BDD\cite{bdd100k} dataset, only SFNet and SFNet-Lite can be compared to our approach. SFNet-Lite\cite{sfnet-lite} and SFNet\cite{sfnet} are currently the state-of-the-art on real-time BDD semantic segmentation. We can observe the experimental results from Table \ref{Bdd10k}. 
\begin{table}[h]
\centering
\setlength{\tabcolsep}{0.05mm}
\renewcommand{\arraystretch}{1.1}
\begin{tabular}{ccccc}
\hline Method & mIOU(\%) & \#FPS & \#Params &GPU \\
\hline \textit{Non-real-time model} & & & & \\
PSPNet\cite{pspnet} & 62.3 &-& $31.1 \mathrm{M}$ &- \\
Deeplabv3+\cite{deeplabv3+} & 63.6 &-& $40.5 \mathrm{M}$ &- \\
DANet\cite{danet} & 62.8 &-& $48.1 \mathrm{M}$ &- \\
OCRNet\cite{ocrnet} & 60.1 &-& $39.0 \mathrm{M}$ &- \\
DSNet-Base & $\mathbf{6 4 . 6}$ &-& $37.5 \mathrm{M}$ &- \\
\hline \textit{Real-time model} & & & & \\
SFNet(DF2)\cite{sfnet} & 60.2 & $\mathbf{2 0 8 . 2}$ & $19.6 \mathrm{M}$ & RTX4090 \\
SFNet(Res18)\cite{sfnet} & 60.6 & 132.5 & $12.9 \mathrm{M}$ & RTX4090 \\
SFNet-Lite(Res18)\cite{sfnet-lite} & 60.6 & 161.3 & $12.3 \mathrm{M}$ & RTX4090 \\
SFNet-Lite(STDC2)\cite{sfnet-lite} & 59.4 & 194.5 & $13.7 \mathrm{M}$ & RTX4090 \\
DSNet-head64 & $\mathbf{6 2 . 8}$ & 172.2 & $\mathbf{6 . 6 M}$ & RTX4090 \\
\hline
\end{tabular}
\caption{Comparison with other state-of-the-art models on BDD. Most of the results can be found in \cite{sfnet-lite}. The FPS is measured at resolution $1280\times720$.}
\label{Bdd10k}
\end{table}
\begin{table*}[h]
\centering
\setlength{\tabcolsep}{0.3mm}
\renewcommand{\arraystretch}{1.1}
\begin{tabular}{cc|c|c|c|c|c|c}
\hline 
Method & Reference & mIOU(\%) &\#FPS(Torch)& GPU & Resolution &\#GFLOPs&\#Params \\ 
\hline 
BiSeNet(Res18)\cite{bisenet} & ECCV2018 & 74.8 & 65.5 & GTX1080Ti & $1536\times 768$ & 55.3 & $49 \mathrm{M}$ \\ 
BiSeNetV2-L\cite{bisenetv2} & IJCV2021 & 75.8 & 47.3 & GTX1080Ti & $1024\times 512$ & 118.5 & - \\ 
SwiftNetRN-18\cite{SwiftNetRN} & CVPR2019 & 75.5 & 39.9 & GTX1080Ti & $2048\times 1024$ & 104.0 & $11.8 \mathrm{M}$ \\ 
STDC-2-Seg75\cite{STDC} & CVPR2021 & 77.0 & 58.2 & RTX3090 & $1536\times768$ & - & $22.2 \mathrm{M}$ \\ 
PP-LiteSeg-T\cite{pp-liteseg} & ArXiv 2022 & 76.0 & 96.0 & RTX3090 & $1536\times768$ & - & - \\ 
PP-LiteSeg-B2\cite{pp-liteseg} & ArXiv 2022 & 78.2 & 68.2 & RTX3090 & $1536\times768$ & - & - \\ 
RegSeg\cite{regseg} & ArXiv 2021 & $78.1 \pm 0.43$ & 30.0 & T4 & $2048\times1024$ & 39.1 & $3.34 \mathrm{M}$ \\ 
SFNet(Res18)*\cite{sfnet} & ECCV2020 & 79.0 & 65.4 & RTX4090 & $2048\times1024$ & 247.0 & $12.9 \mathrm{M}$ \\ 
DDRNet-23*\cite{ddrnet} & T-ITS2022 & $79.1 \pm 0.3$ & $\mathbf{1 3 5 . 1}$ & RTX4090 & $2048\times1024$ & $\mathbf{1 4 3 . 1}$ & $20.1 \mathrm{M}$ \\ 
PIDNet-M*\cite{pidnet} & CVPR2023 & 80.1 & 100.8 & RTX4090 & $2048\times1024$ & 197.4 & $34.4 \mathrm{M}$ \\ 
DSNet-head128* & Ours & $\mathbf{8 0 . 4}$ & 81.9 & RTX4090 & $2048\times1024$ & 226.6 & $6.8 \mathrm{M}$ \\ 
\hline \textit{mmcv-based} & & & & & & \\ 
TopFormer-B-Seg100\cite{topformer} & CVPR 2022b & 76.3 & $\mathbf{8 1 . 4}$ & RTX3090 & $2048\times1024$ & - & $5.1 \mathrm{M}$ \\ 
SeaFormer-B-Seg100\cite{seaformer} & ICLR 2023 & 77.7 & 37.5 & RTX3090 & $2048\times 1024$ & - & $8.6 \mathrm{M}$ \\ 
RTFormer-B\cite{rtformer} & NeurIPS 2022 & 79.3 & 50.2 & RTX3090 & $2048\times1024$ & - & $16.8 \mathrm{M}$ \\ 
AFFormer-B-Seg100\cite{afformer} & AAAI 2023 & 78.7 & 28.4 & RTX3090 & $2048\times1024$ & - & $\mathbf{3 . 0 M}$ \\ 
SegNext-T-Seg100\cite{segnext} & NeurIPS 2022b & 79.8 & 28.1 & RTX3090 & $2048\times 1024$ & - & $4.3 \mathrm{M}$ \\ 
DSNet-head128* & Ours & $\mathbf{8 0 . 4}$ & 37.6 & RTX3090 & $2048\times1024$ & 226.6 & $6.8 \mathrm{M}$ \\ 
\hline 
\end{tabular} 
\caption{Comparisons with other state-of-the-art real-time methods
on Cityscapes Val set. The inference speeds for models marked with * are tested on our platform. All methods measured by single scale inference. The GFLOPs and parameter count of DSNet are calculated using thop.profile.}
\label{CityscapesRealTime}
\end{table*}We acheieved new state-of-the-art on real-time BDD semantic segmentation. Specifically, we achieved an accuracy 2.2\% higher than SFNet (ResNet18) and SFNet-Lite (ResNet18) with higher speed. Compared to faster versions of the SFNet family, we achieved a higher accuracy of 3.4\% mIoU than SFNet-Lite (STDC-2), with a mere 0.3ms decrease in speed. Furthermore, DSNet-Base achieved the highest accuracy in the non-real-time domain. 

\begin{table}[h]
\centering
\setlength{\tabcolsep}{0.2mm}
\renewcommand{\arraystretch}{1.1}

\begin{tabular}{cccc}
\hline Method & backbone & mIOU(\%) & \#Params \\
\hline UNet++\cite{unet++} & ResNet-101 & 75.5 & $59.5 \mathrm{M}$ \\
DeepLabv3\cite{deeplabv3} & D-ResNet-101 & 78.5 & $58.0 \mathrm{M}$ \\
DeepLabv3+\cite{deeplabv3+} & D-Xception-71 & 79.6 & $43.5 \mathrm{M}$ \\
PSPNet\cite{pspnet} & D-ResNet-71 & 79.7 & $65.9 \mathrm{M}$ \\
HRNetV2-W40\cite{hrnet} & HRNetV2-W40 & 80.2 & $45.2 \mathrm{M}$ \\
HRNetV2-W48\cite{hrnet} & HRNetV2-W48 & 81.1 & $65.9 \mathrm{M}$ \\
OCRNet\cite{ocrnet} & HRNetV2-W48 & 81.6 & $70.5 \mathrm{M}$ \\
CCNet\cite{ccnet} & ResNet-101 & 80.2 & - \\
GSCNN\cite{gcsnn} & - & 80.8 & - \\
Ours & DSNet-Base & $\mathbf{82.0}$ & $\mathbf{37.5 M}$ \\
\hline
\end{tabular}
\caption{Comparison of DSNet-Base with other state-of-the-art models on Cityscapes val set.}
\label{Cityscapes}
\end{table}
\noindent \textbf{Cityscapes.} Previous CNN-based real-time works treat Cityscapes as the standard benchmark. As can be observed from Table \ref{CityscapesRealTime}, our method achieves a new state-of-the-art trade-off between real-time and high accuracy. Specifically, DSNet achieves the highest accuracy while maintaining real-time performance. Compared to SeaFormer-B-Seg100, AFFormer-B-Seg100, and SegNext-T-Seg100, DSNet achieves an mIOU accuracy increase of 2.7\%, 1.7\%, and 0.6\%, respectively, while maintaining faster inference speed. When compared to state-of-the-art CNN models such as SFNet, DDRNet23, and PIDNet-M, DSNet still achieves the highest accuracy. From Table \ref{Cityscapes}, furthermore, it can be observed that DSNet-Base remains highly competitive compared to other high-accuracy models such as HRNet and DeepLabV3.

\section{Conclusion}
\hspace{1.5em}This paper revisits the application of atrous convolutions and proposes a new dual-branch network with the same resolution based on several simple guidelines. DSNet achieves promising results on three large datasets. As our method is real-time, applying it can yield significant benefits in practical applications. Furthermore, to further enhance accuracy, one may consider replacing the size of atrous convolutions with $5\times5$ or larger in the context branch, which may lead to a larger receptive field.

{
    \small
    \bibliographystyle{ieeenat_fullname}
    \bibliography{main}
}


\end{document}